\documentclass{article}
\usepackage{ijcai22}

\usepackage{times}
\usepackage{soul}
\usepackage{url}
\usepackage[hidelinks]{hyperref}
\usepackage[utf8]{inputenc}
\usepackage[small]{caption}
\usepackage{graphicx}
\usepackage{float}
\usepackage{amsmath}
\usepackage{multirow}
\usepackage{amsthm}
\usepackage{balance}
\usepackage{booktabs}
\usepackage{algorithm}
\usepackage{algorithmic}

\usepackage{pbox}

\usepackage{graphicx}
\usepackage{amsmath}
\usepackage{amssymb}
\usepackage{booktabs}

\usepackage{caption}
\usepackage{subcaption}
\usepackage{tikz}

\title{Dynamic Group Transformer: A General Vision Transformer Backbone with Dynamic Group Attention}

\author{
Kai Liu$^1$\textsuperscript{\thanks{Equal contribution}\thanks{Interns at the Institute of Deep Learning, Baidu Research}}
\and
Tianyi Wu$^{2,3}$ \textsuperscript{\footnotemark[1]}
\and
Cong Liu$^1$ 
\and
Guodong Guo$^{2,3}$  \textsuperscript{\thanks{Corresponding author}}
\affiliations
$^1$Sun Yat-sen University, Guangzhou, China\\
$^2$Institute of Deep Learning, Baidu Research, Beijing, China\\
$^3$National Engineering Laboratory for Deep Learning Technology and Application, Beijing, China\\
\emails
liuk95@mail2.sysu.edu.cn,
liucong3@mail.sysu.edu.cn,
\{wutianyi01, guoguodong01\}@baidu.com
}

\begin{document}

\maketitle

\begin{abstract}
    Recently, Transformers have shown promising performance in various vision tasks. To reduce the quadratic computation complexity caused by each query attending to all keys/values, various methods have constrained the range of attention within local regions, where each query only attends to keys/values within a hand-crafted window. However, these hand-crafted window partition mechanisms are data-agnostic and ignore their input content, so it is likely that one query maybe attends to irrelevant keys/values. To address this issue, we propose a Dynamic Group Attention (DG-Attention), which dynamically divides all queries into multiple groups and selects the most relevant keys/values for each group. Our DG-Attention can flexibly model more relevant dependencies without any spatial constraint that is used in hand-crafted window based attention. Built on the DG-Attention, we develop a general vision transformer backbone named Dynamic Group Transformer (DGT). Extensive experiments show that our models can outperform the state-of-the-art methods on multiple common vision tasks, including image classification, semantic segmentation, object detection, and instance segmentation.
\end{abstract}
\section{Introduction}

Recently, Transformer has shown a great potential for various vision tasks \cite{vit,touvron2020deit,liu2021swin,cswin}.
The pioneer Vision Transformer \cite{vit} (ViT) stacked multiple Transformer blocks to process non-overlapping image patch (i.e., visual token) sequences for image classification. However, the global self-attention in Transformer makes each query attend to all keys, which has the quadratic complexity to sequence length, and results in expensive computation costs and memory usage, especially for high-resolution images.

\begin{figure}[t]
    \centering
    \includegraphics[width=0.99\columnwidth]{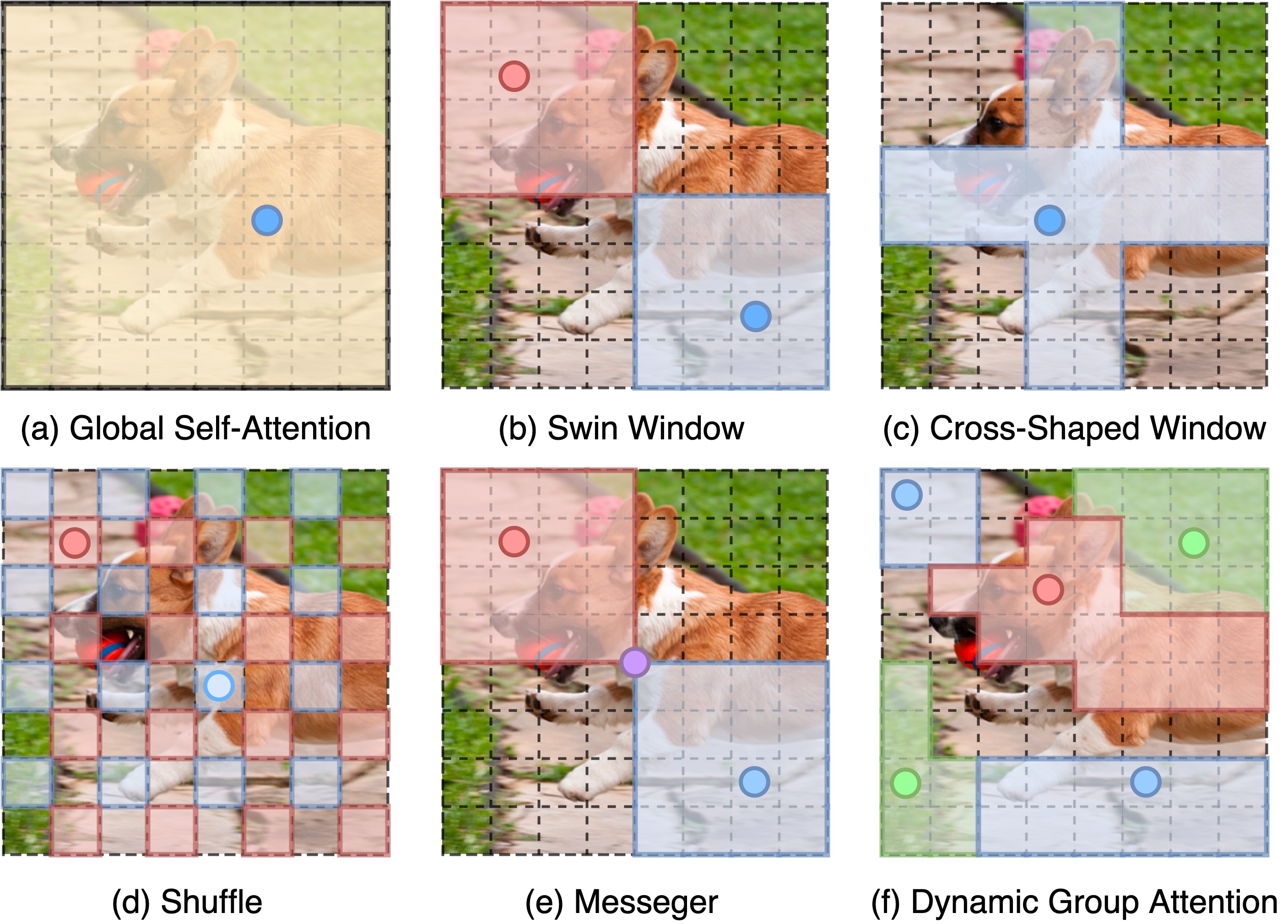}
    \caption{Illustrate our Dynamic Group Attention in comparison with other attention mechanisms in Transformer backbones.
        (a) Global self-attention, each query attends to all keys/values.
        (b) $\sim$ (e) Window-based attentions, each query attends to keys/values within a fixed window.
        (d) Dynamic group attention, all queries are dynamically divided into several groups, and each query attends to relevant keys/values only.
    }
    \label{fig:rel}
\end{figure}

To improve the efficiency of the global self-attention, the state-of-the-art methods \cite{liu2021swin,shuffle,cswin,msg}  focused on how to divide the global image into multiple local regions (or windows). Each query only attends to a few keys within the manually
designed local regions (windows). For example, Swin Transformer \cite{liu2021swin} computed the self-attention within each local window and employed a shifted mechanism to make cross-window connections (Figure \ref{fig:rel} (b)). Different from Swin Transformer,
CSwin \cite{cswin} proposed cross-shaped window self-attention  for computing attention in the horizontal and vertical stripes in parallel that form a cross-shaped window (Figure \ref{fig:rel} (c)).
Shuffle Transformer \cite{shuffle} presented a spatial shuffle operation to make information flow across windows (Figure \ref{fig:rel} (d)).
MSG-Transformer \cite{msg} proposed to compute attention in local regular windows, and used additional MSG tokens to
get connections between them (Figure \ref{fig:rel} (e)).
These window-based methods achieved excellent performances and were superior to the CNN counterparts,
however, they rely on hand-crafted window partition mechanisms.
Furthermore, these partition methods are data-agnostic and ignore the input content,
as a result, it is likely that one query maybe attends to irrelevant keys/values.

To address the issues mentioned above, a good idea is to dynamically select relevant keys/values for each query. However, it leads to unreasonably high memory usage and computation complexity.
We propose dynamic group attention (DG-Attention), which dynamically divides all queries into multiple groups and selects the most relevant keys/values for each group. Specifically, the input visual tokens (or feature vectors) are divided adaptively into multiple groups according to their similarity to all cluster centroids. Therefore, such a partition mechanism is adaptive to input images.
Then, we use the cluster centroid of each group to select the most relevant keys/values subset from the whole keys/values set, and the self-attention is conducted within each group.
It enables our model to focus on relevant keys/values without any spatial constraint.
And also, through the dynamic grouping, our DG-Attention does not cause a high memory usage or large computation cost.
For example, as shown in Figure \ref{fig:rel} (f), the red point (query) can attend to its relevant region denoted with red solid line boundaries, and the blue point can attend to the regions with blue solid line boundaries.
Benefiting from the data-dependent and flexible group mechanism, our DG-Attention shows superiority to the other window-based self-attention illustrated in Figure \ref{fig:rel}.

Based on the proposed DG-Attention, we design a general vision transformer backbone for image classification, named Dynamic Group Transformer (DGT).
We scale our approach up to get a family of models, including DGT-T (24M), DGT-S (52M), and DGT-B (90M).
They achieve significantly a better performance than previous methods.
Our DGT-T can achieve Top-1 classification accuracy of 83.8\% on ImageNet-1k, 50.2\% mIoU on ADE20K for semantic segmentation, 47.7\% box mAP for object detection, and 43.0\% mask mAP on COCO for instance segmentation, outperforming the state-of-the-art methods.
Furthermore, our largest variant DGT-B is also superior to the previous methods, achieving 85.0\% Top-1 accuracy on ImageNet-1K, 51.2\% mIoU on ADE20K, 49.1\% box mAP, and 44.1\% mask mAP on COCO dataset.

\section{Related Work}

This section briefly reviews related works, including improving efficiency and enhancing inductive bias for Vision Transformer.

\noindent \textbf{Improving Efficiency for Vision Transformer.}
There are two main categories of methods to reduce the computation demand for Vision Transformer.
(1) Pruning Token. It aims to remove redundant tokens and reduce the number of tokens entering into attention modules to save the computation demand.
For example, DynamicViT \cite{dynamicvit} pruned tokens in each layer with Gumbel softmax. IA-Red \cite{iared} used reinforcement Learning to achieve a similar effect. Such methods achieved good performances on image classification,
but they are not friendly enough for downstream dense prediction tasks.
(2) Designing efficient attention mechanisms. Such methods mainly explored how to make each query attend to partial keys/values for reducing the computational cost. PVT \cite{wang2021pyramid} directly downsampled the keys and values in each block.
Swin transformer \cite{liu2021swin} divided all queries/keys/values into multiple windows and computed the self-attention within each local window. Similarly, CSwin transformer \cite{cswin} expanded the window into a cross-shaped window. MSG-Transformer \cite{msg} used additional MSG tokens to make connections between windows.
Different from these methods that employed pre-designed, hand-crafted window partition mechanisms, our method dynamically divides
all queries into multiple groups and selects the most relevant keys/values for each group.

\noindent
\textbf{Enhancing Inductive Bias for Vision Transformer.}
Vision transformers have shown successes in various computer vision tasks, due to their ability to
model long-range dependencies within an image. However, recent works also showed that inductive bias could be
incorporated for vision transformers.
CPE \cite{chu2021conditional} used convolution layers to generate the conditional position encoding. CVT \cite{wu2021cvt} employed convolution layers to generate the queries, keys and values. CMT \cite{cmt} also incorporated the convolution layers into the FFN.

\begin{figure*}[t]
    \centering
    \begin{minipage}{\textwidth}
        \centering
        \begin{subfigure}[t]{\textwidth}
            \centering
            \includegraphics[width=0.98\textwidth]{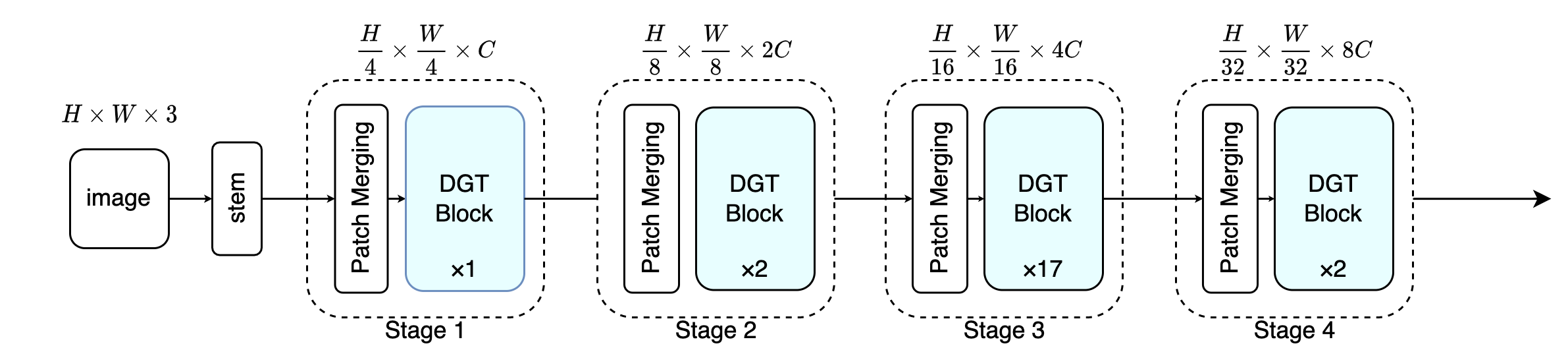}
            \caption{Overall Architecture of Dynamic Group Transformer (DGT)}
        \end{subfigure}
    \end{minipage}

    \begin{minipage}{\textwidth}
        \centering
        \begin{subfigure}[t]{0.24\textwidth}
            \centering
            \includegraphics[width=0.9\textwidth]{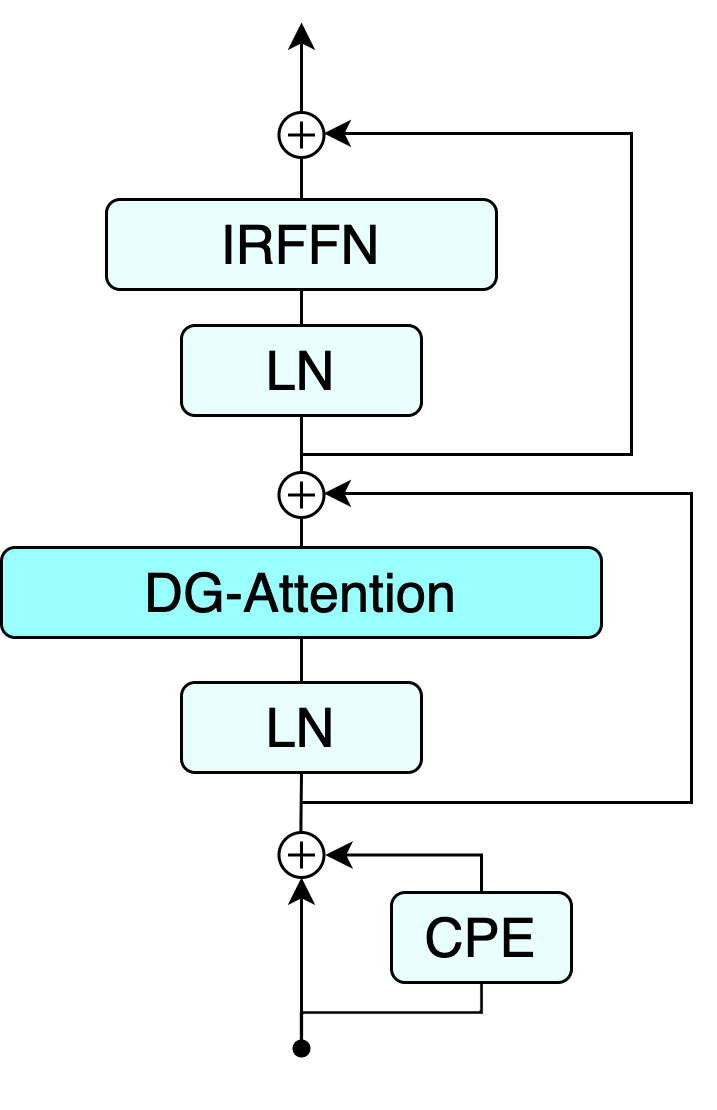}
            \caption{DGT Block}
        \end{subfigure}
        \hfill\vline\hfill
        \begin{subfigure}[t]{0.75\textwidth}
            \centering
            \includegraphics[width=0.9\textwidth]{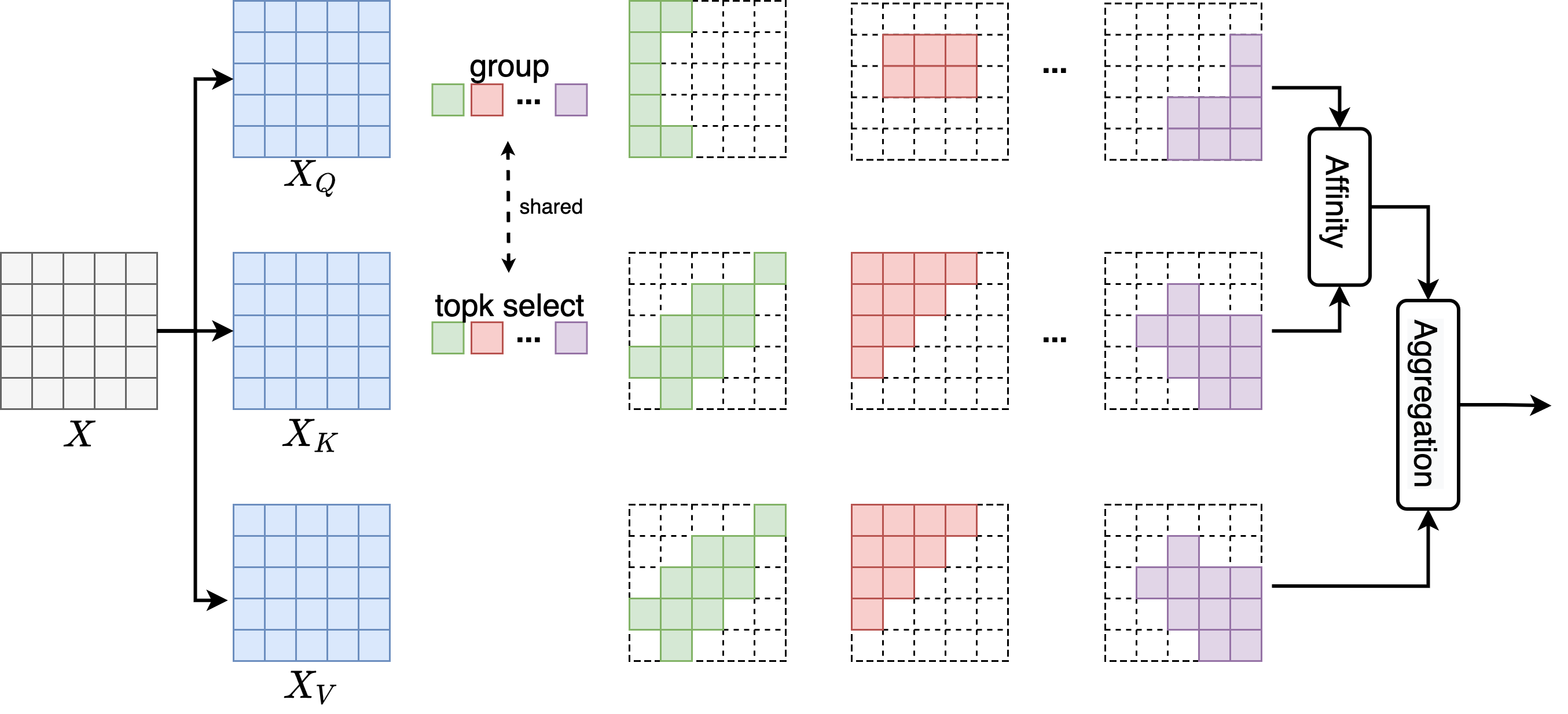}
            \caption{Dynamic Group Attention (DG-Attention)}
        \end{subfigure}
    \end{minipage}
    \caption{(a) The overall architecture of our Dynamic Group Transformer. (b) The composition of each block.
        (c) Illustration of our DG-Attention. It can dynamically divides all queries into multiple groups and selects the most relevant keys/values for each group.}
    \label{fig:method}
\end{figure*}

\section{Method}
In this section, we first introduce our Dynamic Group attention (DG-Attention).
Then, we present the composition of the Dynamic Group Transformer Block.
Finally, we describe the overall architecture and variant configurations of our Dynamic Group Transformer (DGT) backbone.

\subsection{Dynamic Group Attention}

To make each query attend to relevant keys/values, we propose a Dynamic Group Attention (DG-Attention).
It dynamically divides all queries into multiple groups and selects the most relevant keys/values for each group to
compute the self-attention.
As shown in Figure \ref{fig:method}(c), given an input feature map $X\in \mathcal{R}^{H\times W \times C}$ ($C$ is the channel number, $H$ and $W$ denotes the height and width, respectively),
we first get query embeddings $\{X_Q^{i}\}_{i=1}^{L}$ , key embeddings  $\{X_K^{i}\}_{i=1}^{L}$ , and value embeddings  $\{X_V^{i}\}_{i=1}^{L}$, where $L= H\times W$. For simplicity, we assume there is only one head in the DG-Attention. It's easy to expand to the multi-head situation where each head has its queries, keys, values, and cluster centroids.
Then, we use k-means clustering algorithm to dynamically divide all queries into $G$ different query groups (clusters)
$X_Q=\{ X_{Q_j} | X_{Q_j} \in \mathcal{R}^{N_j \times C} \}_{j=1}^{G}$, where $j$ is the group index and $N_j$ is the number of queries in the $j^{th}$ group.
Meanwhile, we use $top\mbox{-} k$ operations to find the $k$ most relevant keys and values for each query group,
which are denoted as $X_K=\{ X_{K_j} | X_{K_j} \in \mathcal{R}^{k \times C} \}_{j=1}^{G}$ and $X_V=\{ X_{V_j} | X_{V_j} \in \mathcal{R}^{k \times C} \}_{j=1}^{G}$, respectively.

Specifically, for the $j^{th}$ query group, we compute the dot product between its cluster centroid $e_j$ and all keys
$ \{X_K^i\}_{i=1}^{L}$, and then select the $k$ most relevant elements according to the dot product sorting,
which can be formulated as follow:

\begin{equation}
    \begin{split}
        & id_j=Top\mbox{-}k (e_j, \{X_K^{i}\}_{i=1}^{L} ) \in \{1,..,L \}^k, \\
        & X_{K_j}=\{ X_K^{i} | i \in id_j  \}   \in \mathcal{R}^{k \times C},                        \\
        & X_{V_j}=\{ X_V^{i} | i \in id_j  \}  \in \mathcal{R}^{k \times C},
    \end{split}
\end{equation}
where $Top\mbox{-}k$ is the function that returns the indices of top $k$ values, and $id_j$ is an index vector.
Then, the self-attention is conducted within each group:
\begin{equation}
    Y_j = SA( X_{Q_j}, X_{K_j}, X_{V_j}) \in \mathcal{R}^{N_j \times C},
\end{equation}
where SA denotes the Self-Attention,
$Y_j$ is the updated output of the $j^{th}$ query group.
Finally, $ \{Y_j\}_{j=1}^{G}$ are scattered into the output $Y \in \mathcal{R}^{L \times C}$ according to their original spatial position indexes.

As each group has a different number of queries, this algorithm cannot be implemented using the general matrix multiplication. We implement this algorithm using CUDA, and the detail can be found in the supplementary material.

To make the training stable, we update the cluster centroids with exponential moving average after each iteration.
Specifically, for the $j^{th}$ cluster centroid, we compute the current cluster centroid  as follow:
\begin{equation}
    e'_j =  \frac{1}{N_j} \sum_{i}  Norm(X_{Q_j}^{i}).
\end{equation}
Then,  we update the cluster centroid as below:
\begin{equation}
    {e_j} =Norm(\tau \times e_j + (1-\tau)\times {e'}_j),
\end{equation}
where $\tau$ is a hyper-parameter to control the update speed.
We empirically set $\tau$ to $0.1 \times lr$, where $lr$ is the learning rate.

\subsubsection{Computation Complexity Analysis}

We analyze the computation complexity of our DG-Attention and the global self-attention to further reveal the efficiency of our method.
Here, we only consider the process of computing the attention maps and weighted sums of values for clarity.
Given input features of size $L \times C$, the global self-attention has a computational complexity of
\begin{equation}
    \Omega_{Global} =  2L^2C.
\end{equation}
In DG-Attention, each query only attends to $k$ keys, so the basic computation complexity of DG-Attention is $2kLC$.
Besides, grouping queries and selecting the most significant $k$ keys require an additional computation cost of $2LGC+kGlogL$.
Therefore, the total computation complexity of our DG-Attention is
\begin{equation}
    \Omega_{DG\mbox{-}Attention} = 2kLC + 2LGC + kGlogL
\end{equation}

The ratio of the computation complexity of our DG-Attention and Global self-attention is:

\begin{equation}
    \begin{split}
        \frac{ \Omega_{DG\mbox{-}Attention} }{\Omega_{Global}} & =\frac{2kLC + 2LGC + kGlogL}{2L^2C} \\
        & = \frac{k}{L} + \frac{G}{L} + \frac{kGlogL}{2L^2C} < 1
    \end{split}
\end{equation}
where $L$ is larger than $G$ and $k$. For high-resolution inputs, the ratio $\frac{ \Omega_{DG\mbox{-}Attention} }{\Omega_{Global}}  <<1 $.
Typically, for the ImageNet classification task,
$k$ is set to 98 for the first three stages, while the corresponding $L$ is 3136, 784, and 196.
Thus, the ratio is 0.05, 0.19, and 0.75 for DGT-T.
Besides, $k$ is independent of the shapes of the parameters in our models,
so we can adjust $k$ to balance the performance and computation efficiency.

\subsection{Dynamic Group Transformer Block}

The Dynamic Group Transformer block is shown in Figure \ref{fig:method}(b).
It first employs the widely-used conditional position embeddings (CPE) \cite{chu2021conditional} to generate the positional information.
Then, DG-Attention is applied to model spatial relevant dependencies flexibly and dynamically.
Last, IRFFN (Inverted Residual Feed-forward Network) \cite{cmt} further is employed to capture local dependencies.
The forward process of the $l^{th}$ block can be formulated as follows:
\begin{align}
     & \tilde{X^l}=X^{l-1}+CPE(X^{l-1}),                             \\
     & \hat{X^l}=\tilde{X^l}+DG\mbox{-}Attention(LN(\tilde{X^{l}})), \\
     & X^l=\hat{X^l}+IRFFN(LN(\hat{X^{l}})),
\end{align}
where $LN(\cdot)$ denotes Layer Normalization, $X^{l}$ and $X^{l-1}$
are the output of the $l^{th}$ and ${l\mbox{-}1} ^{th}$ block, respectively.

\subsection{Overall Architecture and Variants}
Our Dynamic Group Transformer (DGT) consists of a convolutional stem, four hierarchical stages, and a classifier head, as shown in Figure \ref{fig:method} (a).
The stem is designed to extract local dependency, similar to \cite{cmt}, which consists of one 3$\times$3 convolution layer with stride = 2 and two 3$\times$3 convolution layers with stride = 1.
After the stem, each stage contains a patch merging layer and multiple transformer blocks. The first three stages use the DGT block, and the last stage applies global self-attention (GSA) block, which is achieved by replacing the DG-Attention with global self-attention in DGT block.
We decrease the number of tokens and double the channel dimension by using a 3$\times$3 convolutional layer with stride = 2 before each stage to produce a hierarchical representation. The final classifier head consists of two linear layers.

Finally, we design three different variants, including DGT-Tiny (DGT-T), DGT-Small (DGT-S), and DGT-Base (DGT-B), whose detailed configurations are shown in Table \ref{tab:arch}.
For all variants, the number of blocks in each stage is fixed with [1,2,17,2]. In each DGT block, the expand ratios of IRFFN are set to 4, the number of groups $G$ is 48. The number of selected keys/values $k$ is 98 for image classification on ImageNet \cite{Imagenet}.
The main differences among all variants are the channel dimension and the number of heads in DGT blocks. Besides, to train the model stably, we apply post LayerNorm and cosine attention \cite{liu2021swinv2} in DGT-S and DGT-B.

\begin{table}[t]
    \resizebox{1.0\columnwidth}{!}{
        \begin{tabular}{c|c|c|c|c}
            \toprule [0.15em]
            \begin{tabular}{c}Stage/Stride\end{tabular}
               & Layer
               & DGT-T
               & DGT-S
               & DGT-B
            \\
            \hline
            Stride=2
               & Stem
               & $\begin{array}{c} 3\times3, 32, s=2 \\ \left[3\times3, 32\right] \times 2 \end{array}$
               & $\begin{array}{c} 3\times3, 48, s=2\\ \left[3\times3, 48\right] \times 2 \end{array}$
               & $\begin{array}{c} 3\times3, 64, s=2 \\ \left[3\times3, 64\right] \times 2 \end{array}$
            \\
            \hline

            \multirow{ 5}{*}{ \begin{tabular}{c} Stage 1 \\ Stride=4 \end{tabular}}
               & \begin{tabular}{c}Patch \\Merge\end{tabular}
               & $3\times3$, $64$, s=2
               & $3\times3$, $96$, s=2
               & $3\times3$, $128$, s=2
            \\
            \cline{2-5}
               & \begin{tabular}{c}DGT \\Block\end{tabular}
               & $\begin{bmatrix}\setlength{\arraycolsep}{1pt} \begin{array}{c}
                        H_1$=$2  \\
                        G_1$=$48 \\
                        k_1$=$98 \\
                        R_1$=$4
                    \end{array} \end{bmatrix} \times 1$
               & $\begin{bmatrix}\setlength{\arraycolsep}{1pt} \begin{array}{c}
                        H_1$=$3  \\
                        G_1$=$48 \\
                        k_1$=$98 \\
                        R_1$=$4
                    \end{array} \end{bmatrix} \times 1$
               & $\begin{bmatrix}\setlength{\arraycolsep}{1pt} \begin{array}{c}
                        H_1$=$4  \\
                        G_1$=$48 \\
                        k_1$=$98 \\
                        R_1$=$4
                    \end{array} \end{bmatrix} \times 1$
            \\
            \hline
            \multirow{ 5}{*}{ \begin{tabular}{c} Stage 2 \\ Stride=8 \end{tabular}}

               & \begin{tabular}{c}Patch \\Merge\end{tabular}

               & $3\times3$, $128$, s=2
               & $3\times3$, $192$, s=2
               & $3\times3$, $256$, s=2
            \\
            \cline{2-5}
               & \begin{tabular}{c}DGT \\Block\end{tabular}
               & $\begin{bmatrix}\setlength{\arraycolsep}{1pt} \begin{array}{c}
                        H_2$=$4  \\
                        G_2$=$48 \\
                        k_2$=$98 \\
                        R_2$=$4
                    \end{array} \end{bmatrix} \times 2$
               & $\begin{bmatrix}\setlength{\arraycolsep}{1pt} \begin{array}{c}
                        H_2$=$6  \\
                        G_2$=$48 \\
                        k_2$=$98 \\
                        R_2$=$4
                    \end{array} \end{bmatrix} \times 2$
               & $\begin{bmatrix}\setlength{\arraycolsep}{1pt} \begin{array}{c}
                        H_2$=$8  \\
                        G_2$=$48 \\
                        k_2$=$98 \\
                        R_2$=$4
                    \end{array} \end{bmatrix} \times 2$
            \\
            \hline
            \multirow{ 5}{*}{ \begin{tabular}{c} Stage 3 \\ Stride=16 \end{tabular}}

               & \begin{tabular}{c}Patch \\Merge\end{tabular}

               & $3\times3$, $256$, s=2
               & $3\times3$, $384$, s=2
               & $3\times3$, $512$, s=2
            \\

            \cline{2-5}
               & \begin{tabular}{c}DGT \\Block\end{tabular}
               & $\begin{bmatrix}\setlength{\arraycolsep}{1pt} \begin{array}{c}
                        H_3$=$8  \\
                        G_3$=$48 \\
                        k_3$=$98 \\
                        R_3$=$4
                    \end{array} \end{bmatrix} \times 17$
               & $\begin{bmatrix}\setlength{\arraycolsep}{1pt} \begin{array}{c}
                        H_3$=$12 \\
                        G_3$=$48 \\
                        k_3$=$98 \\
                        R_3$=$4
                    \end{array} \end{bmatrix} \times 17$
               & $\begin{bmatrix}\setlength{\arraycolsep}{1pt} \begin{array}{c}
                        H_3$=$16 \\
                        G_3$=$48 \\
                        k_3$=$98 \\
                        R_3$=$4
                    \end{array} \end{bmatrix} \times 17$
            \\

            \hline
            \multirow{3}{*}{ \begin{tabular}{c} Stage 4 \\ Stride=32 \end{tabular}}

               & \begin{tabular}{c}Patch \\Merge\end{tabular}

               & $3\times3$, $512$, s=2
               & $3\times3$, $768$, s=2
               & $3\times3$, $1024$, s=2
            \\

            \cline{2-5}
               & \begin{tabular}{c}GSA \\Block\end{tabular}
               & $\begin{bmatrix}\setlength{\arraycolsep}{1pt} \begin{array}{c}
                        H_4$=$16 \\
                        R_4$=$4
                    \end{array} \end{bmatrix} \times 2$
               & $\begin{bmatrix}\setlength{\arraycolsep}{1pt} \begin{array}{c}
                        H_4$=$24 \\
                        R_4$=$4
                    \end{array} \end{bmatrix} \times 2$
               & $\begin{bmatrix}\setlength{\arraycolsep}{1pt} \begin{array}{c}
                        H_4$=$32 \\
                        R_4$=$4
                    \end{array} \end{bmatrix} \times 2$
            \\

            \hline
            $$ & FC
               & \multicolumn{3}{c}{$1\times1$, $1280$}
            \\

                \hline
            $$ & Classifier
               & \multicolumn{3}{c}{$1\times1$, $1000$}
            \\
            \hline
            \multicolumn{2}{c|}{ Params}
               & $24.09$ M
               & $51.76$ M
               & $90.28$ M
            \\
            \hline
            \multicolumn{2}{c|}{ Flops}
               & $4.35$ G
               & $9.41$ G
               & $16.4$ G
            \\

            \bottomrule[0.15em]
        \end{tabular}
    }
    \caption{Detailed configurations of different variants of our DGT. $H_i$, $G_i$ and $ k_i$ represent the number of heads, group, and the selected key/value in DGT block, respectively. $R_i$ is the expand ratio in IRFFN.}
    \label{tab:arch}

\end{table}

\section{Experiments}

We first compare our Dynamic Group Transformer (DGT) with the state-of-the-art backbones on
ImageNet-1K \cite{Imagenet} for image classification. To further verify the effectiveness and generalization of our backbone,
we perform experiments on ADE20K \cite{ade20k} for semantic segmentation, COCO \cite{coco} for object detection and instance segmentation.
Finally, we analyze the key design of our Dynamic Group Transformer.

\begin{table}[!t]
    \centering
    \resizebox{1.0\columnwidth}{!}{

        \begin{tabular}[t]{l|cc|c}
            \toprule
            Method                             & Param. & FLOPs & Top-1         \\
            \midrule
            DeiT-S~\cite{touvron2020deit}      & 22M    & 4.6G  & 79.8          \\
            PVT-S~\cite{wang2021pyramid}       & 25M    & 3.8G  & 79.8          \\
            Reg-4G~\cite{radosavovic2020reg}   & 21M    & 4.0G  & 80.0          \\
            Swin-T~\cite{liu2021swin}          & 29M    & 4.5G  & 81.3          \\
            CPVT-S~\cite{chu2021conditional}   & 23M    & 4.6G  & 81.5          \\
            CvT-13~\cite{wu2021cvt}            & 20M    & 4.5G  & 81.6          \\
            ViL-S~\cite{zhang2021mvit}         & 25M    & 4.9G  & 82.0          \\
            CSWin-T~ \cite{cswin}              & 23M    & 4.3G  & 82.7          \\
            Eff-B4*~\cite{tan2019efficientnet} & 19M    & 4.2G  & 82.9          \\
            CMT-S~   \cite{cmt}                & 25M    & 4.0G  & 83.5          \\
            \textbf{DGT-T} (ours)              & 24M    & 4.3G  & \textbf{83.8} \\
            \midrule
            PVT-M~\cite{wang2021pyramid}       & 44M    & 6.7G  & 81.2          \\
            PVT-L~\cite{wang2021pyramid}       & 61M    & 9.8G  & 81.7          \\
            Reg-8G~\cite{radosavovic2020reg}   & 39M    & 8.0G  & 81.7          \\

            CvT-21~\cite{wu2021cvt}            & 32M    & 7.1G  & 82.5          \\
            Swin-S~\cite{liu2021swin}          & 50M    & 8.7G  & 83.0          \\
            Twins-B~\cite{chu2021twins}        & 56M    & 8.3G  & 83.2          \\
            ViL-M~\cite{zhang2021mvit}         & 40M    & 8.7G  & 83.3          \\
            CSWin-S~\cite{cswin}               & 35M    & 6.9G  & 83.6          \\
            Eff-B5*~\cite{tan2019efficientnet} & 30M    & 9.9G  & 83.6          \\
            \textbf{DGT-S} (ours)              & 52M    & 9.4G  & \textbf{84.6} \\
            \midrule
            DeiT-B~\cite{touvron2020deit}      & 87M    & 17.5G & 81.8          \\
            CPVT-B~\cite{chu2021conditional}   & 88M    & 17.6G & 82.3          \\
            Reg-16G~\cite{radosavovic2020reg}  & 84M    & 16.0G & 82.9          \\
            ViL-B~\cite{zhang2021mvit}         & 56M    & 13.4G & 83.2          \\
            Swin-B~\cite{liu2021swin}          & 88M    & 15.4G & 83.3          \\
            Twins-L~\cite{chu2021twins}        & 99M    & 14.8G & 83.7          \\
            Eff-B6*~\cite{tan2019efficientnet} & 43M    & 19.0G & 84.0          \\
            CSWin-B~\cite{cswin}               & 78M    & 15.0G & 84.2          \\
            \textbf{DGT-B} (ours)              & 90M    & 16.4G & \textbf{85.0} \\
            \bottomrule
        \end{tabular}
    }
    \caption{Comparison with the state-of-the-art models, trained with 224$\times$ 224 on ImageNet-1K Classification.}
    \label{tab:imagenet}
\end{table}

\subsection{Image Classification on ImageNet-1K}

We conduct experiments on ImageNet-1K \cite{Imagenet} dataset, which has 1.28M images in the training set and 50K images in the validation set.
Detailed settings are described in the supplementary material.

\subsubsection{Results}

Table \ref{tab:imagenet} compares the performance of our DGT with the state-of-the-art CNN models and vision transformer backbones on ImageNet-1K validation set.
We can see that our DGT variants outperform the state-of-the-art CNN models and vision transformer models when using similar FLOPs.
DGT-T achieves 83.8\% top-1 accuracy with 4.3G FLOPs and outperforms the CNN models Reg-4G and Efficient B4 by 3.8\% and 0.9\%, respectively.
Meanwhile, our DGT outperforms the advanced Transformer-based backbones, and is $+1.6\%$ and $+0.8\%$ higher than Swin and CSwin Transformer, respectively, for all variants under the similar model size and FLOPs. For example, our DGT-T can surpass PVT-S, Swin-T, and CSwin-S by 4.0\%, 2.5\%, and 1.1\%, respectively.
Our DGT-B can outperform Swin-B and CSwin-B by $1.7\%$ and $0.8\%$, respectively.
These results demonstrate the effectiveness and efficiency of our approach.

\subsection{Semantic Segmentation on ADE20K}

To demonstrate the superiority of our Dynamic Group Transformer for semantic segmentation.
We conduct experiments on ADE20K with the widely-used UperNet \cite{xiao2018unified} framework for fair comparisons to other backbones.
Detailed implementation can be found in the supplementary material.

\begin{table}[t]
    \centering
    \resizebox{\columnwidth}{!}{
        \begin{tabular}{l|*{2}c|c}
            \toprule
            Backbone                         & Prams (M) & FLOPs (G) & \begin{tabular}[c]{@{}c@{}} mIoU \\ SS/MS\end{tabular} \\
            \midrule
            TwinsP-S~\cite{chu2021twins}     & 54.6      & 919       & 46.2/47.5                                              \\
            Twins-S~\cite{chu2021twins}      & 54.4      & 901       & 46.2/47.1                                              \\
            Swin-T~\cite{liu2021swin}        & 59.9      & 945       & 44.5/45.8                                              \\
            CrossFormer-S~\cite{crossformer} & 62.3      & 980       & 47.6/48.4                                              \\
            \textbf{DGT-T} (ours)            & 52.5      & 954       & \textbf{50.2/50.8}                                     \\
            \midrule
            Res101~\cite{he2016deep}         & 86.0      & 1029      & --/44.9                                                \\
            TwinsP-B~\cite{chu2021twins}     & 74.3      & 977       & 47.1/48.4                                              \\
            Twins-B~\cite{chu2021twins}      & 88.5      & 1020      & 47.7/48.9                                              \\
            Swin-S~\cite{liu2021swin}        & 81.3      & 1038      & 47.6/49.5                                              \\
            CrossFormer-B~\cite{crossformer} & 83.6      & 1090      & 49.7/50.6                                              \\
            \textbf{DGT-S} (ours)            & 81.9      & 1074      & \textbf{50.8/51.6}                                     \\
            \midrule
            TwinsP-L~\cite{chu2021twins}     & 91.5      & 1041      & 48.6/49.8                                              \\
            Twins-L~\cite{chu2021twins}      & 133.0     & 1164      & 48.8/50.2                                              \\
            Swin-B~\cite{liu2021swin}        & 121.0     & 1188      & 48.1/49.7                                              \\
            CrossFormer-L~\cite{crossformer} & 125.5     & 1244      & 50.5/51.4                                              \\
            \textbf{DGT-B} (ours)            & 122.2     & 1234      & \textbf{51.2/51.8}                                     \\
            \bottomrule
        \end{tabular}
    }
    \caption{Comparison with different backbones on ADE20K. FLOPs are calculated with the resolution of 512$\times $2048.}
    \label{tab:ade20k}
\end{table}

\subsubsection{Results}
Table \ref{tab:ade20k} shows the comparisons of UperNet \cite{xiao2018unified} with various advanced Transformer backbones on ADE20K validation set.
We report both single-scale (SS) mIoU and multi-scale (MS) mIoU for a cleaner comparison.
Our DGT variants outperforms the state-of-the-art methods consistently.
Specifically, our DGT-T achieves 50.2\% mIoU with single scale testing, outperforming the Swin-T and CrossFormer-S by 5.7\% and 2.6\%.
Our DGT-S outperforms Swin-S and CrossFormer-B by 3.2\% and 0.9\% SS mIoU.
Besides, our DGT-B achieves 51.2\%/51.8\% SS/MS mIoU, outperforming the Swin-B and CrossFormer-L by 3.1\%/2.1\% and 0.7\%/0.4\%.
These results demonstrate the advantages of our Dynamic Group Transformer for semantic segmentation.

\subsection{Object Detection and Instance Segmentation on COCO}

We further evaluate our DGT backbone on COCO \cite{coco} dataset for object detection and instance segmentation.
Following previous works \cite{liu2021swin,cswin}, we utilize Mask R-CNN \cite{he2017mask} framework under 1x schedule.
More details are provided in the supplementary material.

\begin{figure*}[t]
    \centering
    \includegraphics[width=\textwidth]{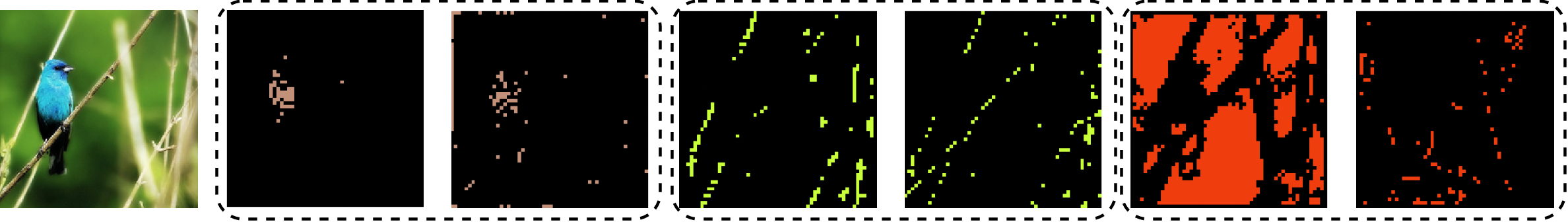}
    \caption{Visualizations of some query and keys groups. Each dashed box contains a query group and its corresponding keys group.  It can be seen that our method can flexibly model relevant dependencies without any spatial constraint. }
    \label{fig:visual_2}
\end{figure*}

\begin{table*}[t]
    \centering
    \begin{minipage}{0.65\textwidth}
        \centering
        \begin{subfigure}[t]{\textwidth}{
                \centering
                \resizebox{\textwidth}{!}{
                    \begin{tabular}{l|*{2}c|*{6}c}
                        \toprule
                        Backbone                         & Params (M) & FLOPS (G) & AP$^{\rm b}$  & AP$_{50}^{\rm b}$ & AP$_{75}^{\rm b}$ & AP$^{\rm m}$  & AP$_{50}^{\rm m}$ & AP$_{75}^{\rm m}$ \\
                        \toprule
                        Res50~\cite{he2016deep}          & 44         & 260       & 38.0          & 58.6              & 41.4              & 34.4          & 55.1              & 36.7              \\
                        PVT-S~\cite{wang2021pyramid}     & 44         & 245       & 40.4          & 62.9              & 43.8              & 37.8          & 60.1              & 40.3              \\
                        ViL-S~\cite{zhang2021mvit}       & 45         & 218       & 44.9          & 67.1              & 49.3              & 41.0          & 64.2              & 44.1              \\
                        TwinsP-S~\cite{chu2021twins}     & 44         & 245       & 42.9          & 65.8              & 47.1              & 40.0          & 62.7              & 42.9              \\
                        Twins-S~\cite{chu2021twins}      & 44         & 228       & 43.4          & 66.0              & 47.3              & 40.3          & 63.2              & 43.4              \\
                        Swin-T~\cite{liu2021swin}        & 48         & 264       & 42.2          & 64.6              & 46.2              & 39.1          & 61.6              & 42.0              \\
                        CrossFormer-S~\cite{crossformer} & 50         & 301       & 45.4          & 68.0              & 49.7              & 41.4          & 64.8              & 44.6              \\
                        \textbf{DGT-T} (ours)            & 42         & 272       & \textbf{47.7} & \textbf{69.8}     & \textbf{52.2}     & \textbf{43.0} & \textbf{66.9}     & \textbf{46.4}     \\
                        \midrule
                        Res101~\cite{he2016deep}         & 63         & 336       & 40.4          & 61.1              & 44.2              & 36.4          & 57.7              & 38.8              \\
                        X101-32~\cite{xie2017resx}       & 63         & 340       & 41.9          & 62.5              & 45.9              & 37.5          & 59.4              & 40.2              \\
                        PVT-M~\cite{wang2021pyramid}     & 64         & 302       & 42.0          & 64.4              & 45.6              & 39.0          & 61.6              & 42.1              \\
                        ViL-M~\cite{zhang2021mvit}       & 60         & 261       & 43.4          & ----              & ----              & 39.7          & ----              & ----              \\
                        TwinsP-B~\cite{chu2021twins}     & 64         & 302       & 44.6          & 66.7              & 48.9              & 40.9          & 63.8              & 44.2              \\
                        Twins-B~\cite{chu2021twins}      & 76         & 340       & 45.2          & 67.6              & 49.3              & 41.5          & 64.5              & 44.8              \\
                        Swin-S~\cite{liu2021swin}        & 69         & 354       & 44.8          & 66.6              & 48.9              & 40.9          & 63.4              & 44.2              \\
                        CrossFormer-B~\cite{crossformer} & 72         & 408       & 47.2          & 69.9              & 51.8              & 42.7          & 66.6              & 46.2              \\
                        \textbf{DGT-S} (ours)            & 70         & 386       & \textbf{48.4} & \textbf{70.7}     & \textbf{53.2}     & \textbf{43.5} & \textbf{67.6  }   & \textbf{47.0 }    \\
                        \midrule
                        X101-64~\cite{xie2017resx}       & 101        & 493       & 42.8          & 63.8              & 47.3              & 38.4          & 60.6              & 41.3              \\
                        PVT-L~\cite{wang2021pyramid}     & 81         & 364       & 42.9          & 65.0              & 46.6              & 39.5          & 61.9              & 42.5              \\
                        ViL-B~\cite{zhang2021mvit}       & 76         & 365       & 45.1          & ----              & ----              & 41.0          & ----              & ----              \\
                        TwinsP-L~\cite{chu2021twins}     & 81         & 364       & 45.4          & ----              & ----              & 41.5          & ----              & ----              \\
                        Twins-L~\cite{chu2021twins}      & 111        & 474       & 45.9          & ----              & ----              & 41.6          & ----              & ----              \\
                        Swin-B~\cite{liu2021swin}        & 107        & 496       & 46.9          & ----              & ----              & 42.3          & ----              & ----              \\
                        \textbf{DGT-B} (ours)            & 108        & 540       & \textbf{49.1} & \textbf{70.9}     & \textbf{54.1}     & \textbf{44.1} & \textbf{68.1}     & \textbf{47.6}     \\
                        \bottomrule
                    \end{tabular}
                }
            }
            \caption{Comparision with different backbones on COCO. Flops are calculated with the resolution of 800$\times$1280.}
        \end{subfigure}
    \end{minipage}
    \hspace{0.3cm}
    \begin{minipage}{0.2\textwidth}
        \centering
        \begin{subfigure}[t]{\textwidth}{
                \centering
                \resizebox{0.7\textwidth}{!}{
                    \begin{tabular}{cc|c}
                        \toprule
                        $G$ & $k$ & Top-1         \\
                        \midrule
                        24  & 98  & 83.6          \\
                        48  & 49  & 83.7          \\
                        48  & 98  & \textbf{83.8} \\
                        \bottomrule
                    \end{tabular}}
                \caption{Effect of Hyper-parameters $G$ and $k$.}
            }
        \end{subfigure}
        \begin{subfigure}[t]{\textwidth}{
                \centering
                \vspace{0.5cm}
                \resizebox{\textwidth}{!}{
                    \begin{tabular}{l|c}
                        \toprule
                        Block            & Top-1         \\
                        \midrule
                        Swin block       & 82.8          \\
                        CSwin block      & 83.0          \\
                        CMT block        & 83.4          \\
                        DGT block (ours) & \textbf{83.8} \\
                        \bottomrule
                    \end{tabular}}
                \caption{Comparison with different Vision Transformer Blocks.}
            }
        \end{subfigure}
    \end{minipage}
    \caption{Experiments on COCO and ablation studies.}
    \label{tab:detection}
\end{table*}

\subsubsection{Results}

The results on COCO dataset are shown in Table \ref{tab:detection}(a).
All DGT variants outperform the state-of-the-art vision transformer backbones under similar FLOPs.
Specifically, for object detection, our DGT-T, DGT-S, and DGT-B can achieve 47.7\%, 48.4\% and 49.1\%  box AP, which surpass Swin by 5.5\%, 3.6\%, and 2.2\%, respectively.
For instance segmentation, our DGT-T, DGT-S, and DGT-B are 3.9\%, 2.6\%, and 1.8\% mask AP higher than the Swin.
Besides, DGT-T outperforms CrossFormer-S by 2.3\% bos AP on object detection and 1.6\% mask AP on instance segmentation. DGT-S outperforms CrossFormer-B by 1.2\% box AP on object detection and 0.8\% mask AP on instance segmentation.

\subsection{Ablation Studies}

We conduct ablation studies for the key designs of our methods on the image classification task.
All experiments are performed with the Tiny variant under the same training settings.

\subsubsection{Effect of Hyper-parameters $G$ and $k$}

First, we validate the effect of the hyper-parameters $G$ and $k$. $G$ is the number of groups. A small $G$ causes the queries in a group to be very different, so the selected keys cannot suit all queries. $k$ determines how many keys each query attends to. There will be too much information loss if $k$ is too small.
The default setting of our model on ImageNet is $G=48$ and $k=98$. We compare the default setting with halving $G$ from 48 to 24 and halving $k$ from 98 to 49.

The results are shown in Table \ref{tab:detection}(b).
Decreasing $G$ or $k$ leads to a poorer performance. Halving $G$ and $k$ decrease the top-1 accuracy by 0.2\% and 0.1\%, respectively. We can balance the performance and efficiency by adjusting $G$ and $k$.

\subsubsection{Comparison with Related Transformer Blocks}

To validate the design of our DGT block, which uses convolution layers to extract local dependency and uses the DG-Attention to extract non-local dependency, we replace our DGT block with other blocks and compare their performance.
We select three blocks: Swin block, CSwin block, and CMT block.
Swin block and Cswin block use shifted window-based self-attention and cross-shaped window self-attention, respectively.
The results are shown in Figure \ref{tab:detection}(c). Our DGT block obviously outperforms Swin block, CSwin block and
CMT block by 1.0\%, 0.8\% and 0.4\%.

\section{Visualization}
We visualize the query groups and their corresponding selected keys with an example shown in Figure \ref{fig:visual_2}.
One can find : 1) different queries prefer to attends to different keys according to their content. These three groups mainly contain the queries of the bird, branches, and background, and they also attend to the keys of the bird, branches, and background, respectively. 2) A query may prefer to attend to a long-range area rather than a short-range local region. These findings show the advantages of our DG-Attention. More visualization examples can be found in the supplementary material.

\section{Conclusion}

We have presented an effective dynamic attention mechanism named Dynamic Group Attention (DG-Attention), which dynamically divides input queries into multiple groups and selects relevant keys/values for each group.
DG-Attention can model more relevant context dependencies than the previous pre-designed window-based local attention mechanism.
Based on the proposed DG-Attention, we have developed a general Vision Transformer backbone, Dynamic Group Transformer (DGT), which can outperform the state-of-the-art on ImageNet-1K for image classification.
Furthermore, our DGT outperforms the existing Vision Transformer backbones on ADE20K for semantic
segmentation, and COCO for object detection and instance segmentation.

\appendix

\section*{Appendix}

This appendix provides the detailed experimental settings for classification, semantic segmentation, object detection, and instance segmentation, respectively. Then, we detail the implementation of DG-Attention. Last, we provide more visualization examples on the last page.

\section{Detailed Experiments Setting}

For a fair comparison, we follow the experiment settings of previous work \cite{liu2021swin} on all datasets.

\subsubsection{Image Classification on ImageNet-1K}

On ImageNet-1k, we train our model for 300 epochs with the AdamW \cite{adamw} optimizer.
The weight decay is set to 0.05 for DGT-Tiny and DGT-Small, and 0.1 for DGT-Base.
The default batch size is 1024, and the initial learning rate is 0.001.
We use cosine learning rate decay and linear warm-up for 20 epochs. Data augmentations include increasing stochastic depth \cite{stochasticdepth}(0.2 (DGT-Tiny), 0.4 (DGT-Small), and 0.5 (DGT-Base)), rand-augment \cite{randaug}, mixup \cite{mixup}, cutmix \cite{cutmix}, random erasing \cite{randerasing}, and  Exponential Moving Average(EMA)\cite{ema}. During the evaluation, the images are first resized to $ 256 \times 256$ and then
center-cropped to $224 \times 224$.

\subsubsection{Semantic Segmentation on ADE20K}

On ADE20k, we use AdamW optimizer, and the weight decay is set to 0.01 for DGT-T and DGT-S, and 0.06 for DGT-B. The initial learning rate is set to 6e-5 with learning rate warmups with 1500 iterations and a linear learning rate decay strategy. We train the model for 160K iterations with batch size of 16 and crop size of 512 × 512. The data augmentations include random horizontal flipping, random re-scaling, and random photo-metric distortion.
Besides, the number of the selected keys $k$ is set to 512 in our DGT backbones for ADE20k.
We report both the single-scale (SS)  mIoU and multi-scale (MS) mIoU of all models.

\subsubsection{Object Detection \& Instance Segmentation on COCO}

On COCO, we train our models for 1x schedule with batch size of 16.
AdamW \cite{adamw} optimizer is used with weight decay of 0.05 for DGT-T and DGT-S, 0.1 for DGT-B. The learning rate is set to 0.0001 and declines at the 8th and 11th epoch with decay rate 0.1.
The stochastic depth rate is set to 0.2, 0.4, and 0.5 for DGT-Tiny, DGT-Small, and DGT-Base, respectively.
Besides, the number of the selected keys $k$ is set to 500 in our DGT backbones for COCO.

\section{Additional Attention Mechanism Comparison}

To further evaluate the efficiency of our DG-Attention, we replace the attention module and position encoding in Swin-T and Cswin-T with our DG-attention and CPE encoding. DG-Attention increases the top-1 accuracy by 1.2\% and 0.7\% for Swin-T and Cswin-T, respectively, as shown in Table \ref{tab:attn}.

\begin{table}[t]
    \centering
    \begin{tabular}[]{c|c}
        \toprule
        Model                & Top-1      \\
        \midrule
        Swin-T               & 81.3       \\
        Swin-T+DG-Attention  & 82.5(+1.2) \\
        \midrule
        Cswin-T              & 82.7       \\
        Cswin-T+DG-Attention & 83.4(+0.7) \\
        \bottomrule
    \end{tabular}
    \caption{Replacing the attention modules in Swin-T and Cswin-T with DG-Attention. For Swin-T, We set M=49 to keep the same number of keys each query attends to for a fair comparision.}
    \label{tab:attn}
\end{table}

\section{Implementation of DG-Attention}

As dynamic group attention(DG-Attention) cannot be implemented with standard matrix multiplication, we implement a group matrix multiplication using CUDA to implement DG-Attention.
\subsubsection{Revisit Dynamic Group Attention}
For simplicity, we define the inputs of dynamic group attention are queries, keys, and values. They are denoted by $X_Q\in R^{L\times C}, X_K\in R^{L\times C}, X_V\in R^{L\times C}$, $L$ is the number of tokens, and C is the dimension of the head. The output is the weighted sum of values, denoted by $Y\in R^{L\times C}$

All queries are divided into $G$ groups using the cluster algorithm in DG-Attention. Different groups may contain a different number of queries. We define $i^{th}$ group have $N_i$ quires. We concat all queries in the group, denoted by $X_{Q_i} \in R^{N_i\times C}$. Its corresponding output is $Y_i\in R^{N_i\times C}$

For each group, we need to select top$k$ keys. $k$ is a constant for all groups. We use $id\in R^{G\times k}$ to denote the index of selected keys. Take $i^{th}$ group as an example, we cat all selected keys according to index $id_i\in R^{1\times M}$ and get $X_{K_i}\in R^{k\times C}$. We also cat their corresponding values and get $X_{V_i}\in R^{k\times C}$.

For $i^{th}$ group, we will compute the following equation:
\begin{align}
    \label{eq:prob1}
     & P_i=X_{Q_i} \times X_{K_i}^T,    \\
     & \hat{P_i}=softmax(P_i)/\sqrt{C}, \\
     & Y_i=P_i \times X_{V_i},
    \label{eq:prob2}
\end{align}
where $P_i\in R^{N_i \times k}$. $P_i$ is the unormalized attention weight of $X_{Q_i}$, the queries in $i^{th}$ group. It's a part of $P \in R^{L\times C}$, the unormalized attention weight for all queries.

Besides, we need to compute the gradient of the input matrix of equation \ref{eq:prob1} and \ref{eq:prob2} as follows:

\begin{align}
    \label{eq:prob3}
     & Grad_{X_{Q_i}}=Grad_{P_i}\times (X_{K_i}^T)^T, \\
    \label{eq:prob4}
     & Grad_{X_{K_i}^T}=X_{Q_i}^T\times Grad_{P_i},   \\
    \label{eq:prob5}
     & Grad_{P_i}=Grad_{Y_i} \times V_i^T,            \\
    \label{eq:prob6}
     & Grad_{X_{V_i}}=P_i^T \times Grad_{Y_i},
\end{align}
where $Grad_{variable}$ indicates the gradient of $variable$. $Y_i \in R^{N_i \times C}$ is the weighted sums of values for $X_{Q_i}$. $Y_i$ is a part of $Y\in R^{L\times C}$, which is the weighted sum of values for all queries.

Computing one group is easy, but as each group has different numbers of queries. Therefore, it is difficult to compute all groups at the same time using standard matrix multiplication with CUDA. Refer to standard CUDA implementation of matrix multiplication\footnote{https://github.com/NVIDIA/cutlass}, we implement group-wise matrix multiplication using CUDA. The group-wise matrix multiplication has four forms. The form 1, 2, 3, 4 are used to solve the group-wise version of the equation \ref{eq:prob1} and \ref{eq:prob5}, \ref{eq:prob2} and \ref{eq:prob3}, \ref{eq:prob4}, \ref{eq:prob6}, respectively.

\begin{figure}[t]
    \centering
    \includegraphics[width=0.5\textwidth]{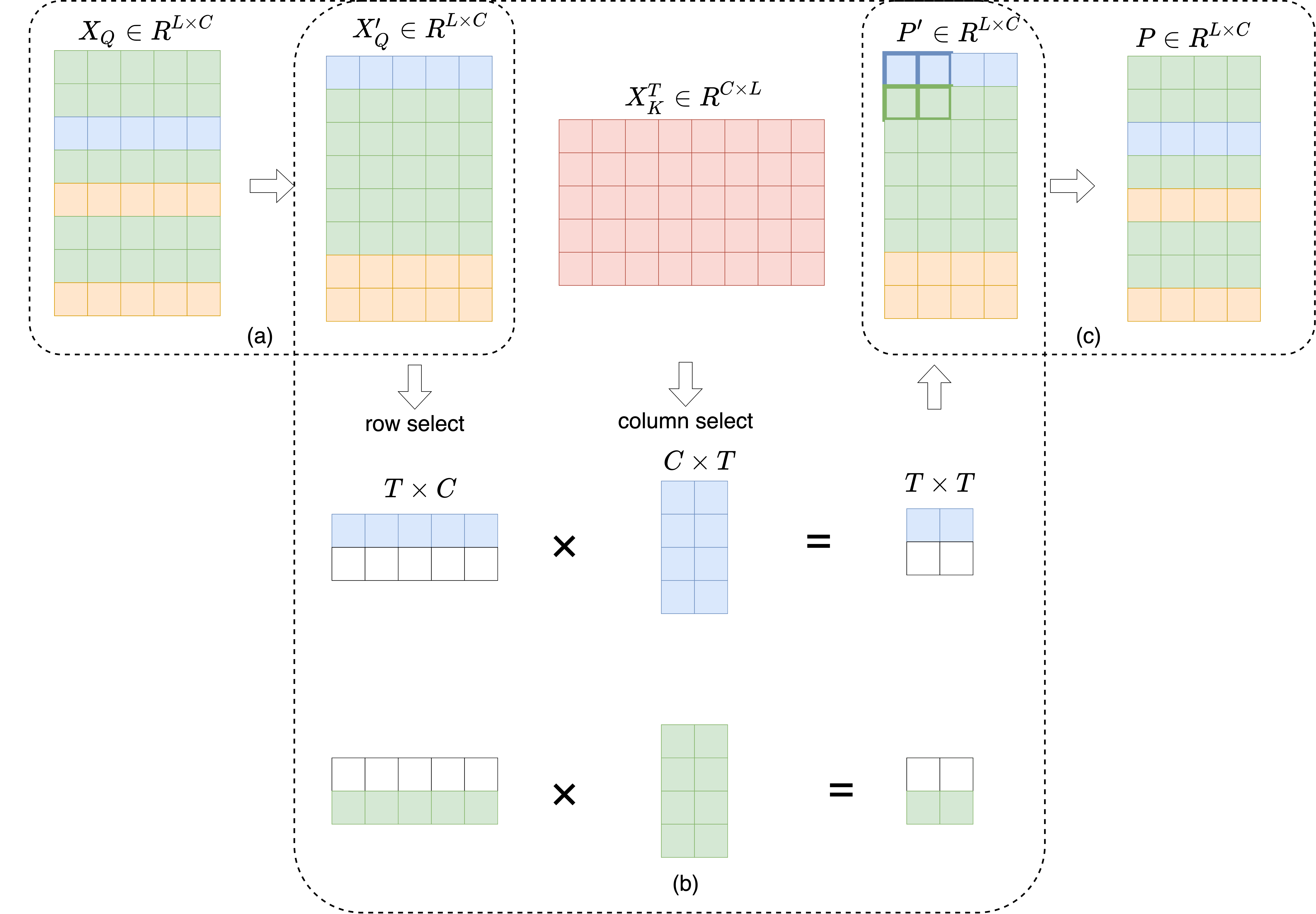}
    \caption{The illustration of computing Equation \ref{eq:prob1}}
    \label{fig:prob}
\end{figure}

\subsection{Form 1}

The Form 1 is used to solve the group-wise version of equation \ref{eq:prob1} and \ref{eq:prob5}.
Take the equation \ref{eq:prob1} as example.
The whole process of computing the group-wise version of equation \ref{eq:prob1} is shown in Figure \ref{fig:prob}.
We solve it in 3 steps.

First, we sort the matrix $X_Q$ according to which group each query belongs. We put the queries in a group to a continuous space and get $X_Q'$, $X_Q'=[X_{Q_1},X_{Q_2},...,X_{Q_G}]$. This process is shown in Figure \ref{fig:prob}(a).

Then, we use a CUDA block to compute a tile of the result matrix $P\in R^{L\times k}$. and each tile has a shape of $T\times T$. In the process, we need to load the corresponding rows of $X_Q$ and columns of $X_K^T$ into the shared memory.
Compared with standard matrix multiplication, our implementation has two key differences. 1), we do not generate $X_{K_i}$ for each group in the global memory to save the global memory resource. We directly pass the matrix $X_K^T$ into the CUDA kernel and extract related columns of $X_{K_i}^T$ into the shared memory for computing. This change lets us save much global memory when increasing the size of $G$ and $M$.
2), For $X_Q'$, a tile may be related to several groups which need different $X_{K_i}$ loaded. To address this problem, we solve different groups in order. The unrelated rows of $Q'$ will be set to zero in the shared memory. When writing to the global memory, the masked part also will be ignored.
The process is shown as Figure \ref{fig:prob}(b).

Last, after completing the CUDA kernel, we get $P'\in R^{L\times C}$. We sort rows of $P'$ into their corresponding position as in $X_Q$ and get the result $P$.

\subsection{Form 2}
Form 2 is used to solve the group-wise version of equation \ref{eq:prob2} and \ref{eq:prob3}.
It is similar to Form 1. Take the equation \ref{eq:prob2} as example.

We first sort $P$ according to the assignment of quires and get $P'$.
Then, we compute the result using CUDA kernel. The only difference with Form 1 in this step is that we select the rows in $X_{V_i}$ in equation \ref{eq:prob2} rather than the columns in $X_{K}^T$ in equation \ref{eq:prob1}.
Last, we sort $Y'$, the output of CUDA kernel, to get the output $Y$.

\begin{figure}[t]
    \centering
    \includegraphics[width=0.5\textwidth]{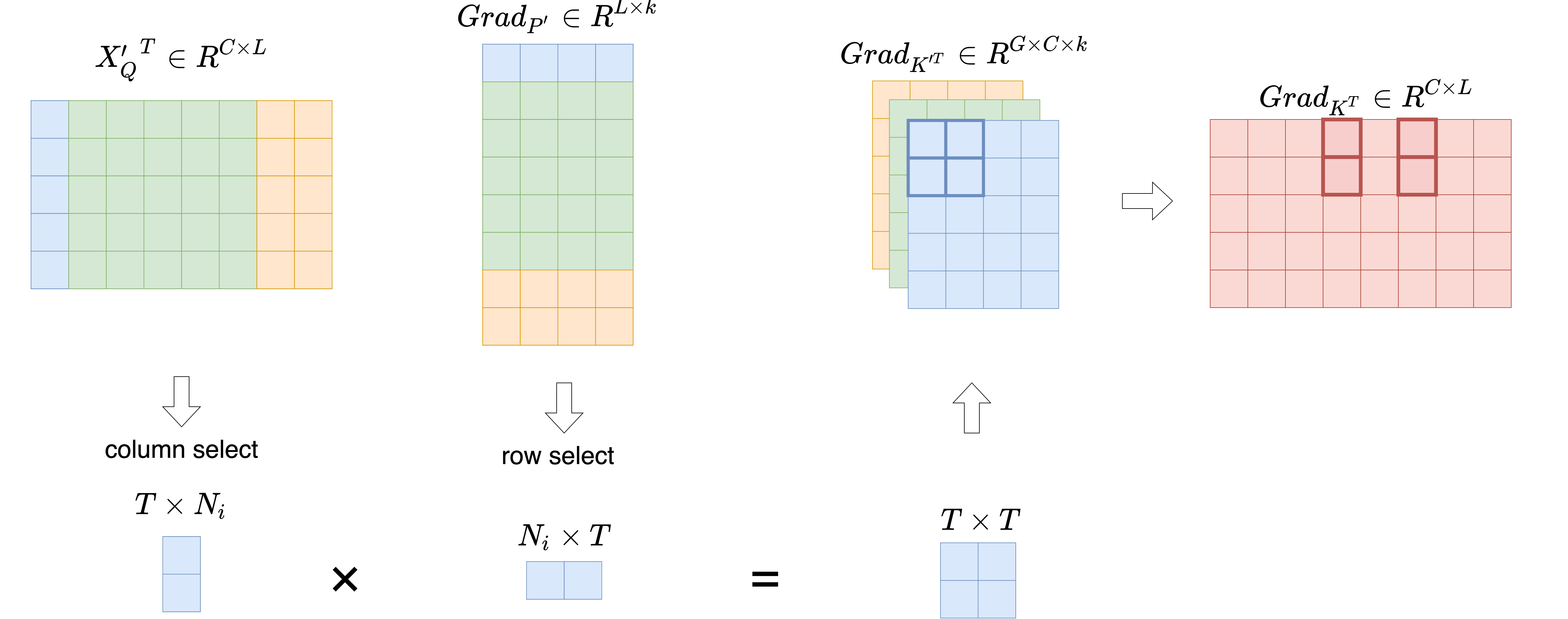}
    \caption{The illustration of computing equation \ref{eq:prob4}}
    \label{fig:back}
\end{figure}

\subsection{Form 3}

The Form 3 is used to solve the group-wise version of equation \ref{eq:prob4}. The possess is shown in Figure \ref{fig:back}.
We concat all $X_{K_i}$ and get $X_{K}'\in R^{G\times k \times C}$. $Grad_{X_{K_i}^T}$ is got by gathering $Grad_{X_K'^T}$ corresponding to the key selection position. We assign CUDA blocks according to $Grad_{X_{K'^T}}$ rather than $Grad_{X_{K^T}}$ for simplicity, where  a CUDA block is used to compute a tile of $Grad_{X_{K_i^T}} \in R^{C \times k}$ .
For each tile, we load the corresponding columns in $X_Q^T$ and rows in $Grad_{P_i}$ into the shared memory to compute the results. However, we do not assign global memory for $Grad_{X_{K'^T}}$ to save the global memory. The columns of $Grad{K'^T}$ are directly written to $Grad_{K^T}$ according to the position of the selected keys in the group.

\subsection{Form 4}

The Form 4 is used to solve equation \ref{eq:prob6}. Form 4 is similar to Form 3. We concat all $X_{V_i}$ and denote it by $X_{V_i}'\in R^{G \times k \times C}$. We assign CUDA blocks according to $Grad_{X_{V}'}$ rather than $Grad_{X_{V}}$. We also do not assign the global memory for $Grad_{V'}$, and directly write the results to $Grad_{V}$. The only difference between Form 3 and Form 4 is that the columns of $Grad_{K'^T}$ are written back $Grad_{K^T}$ in Form 3 and the rows of $Grad_{X_{V}'}$ are written back $Grad_{X_{V}}$ according to the position of the selected values.

With the four forms of group-wise matrix multiplication, we can solve the equation 1-7. We will open-source our implementation after the paper is accepted.

\section{Visualization}

\begin{figure*}[t]
    \includegraphics[width=\textwidth,trim={0 5.4cm 0 0},clip]{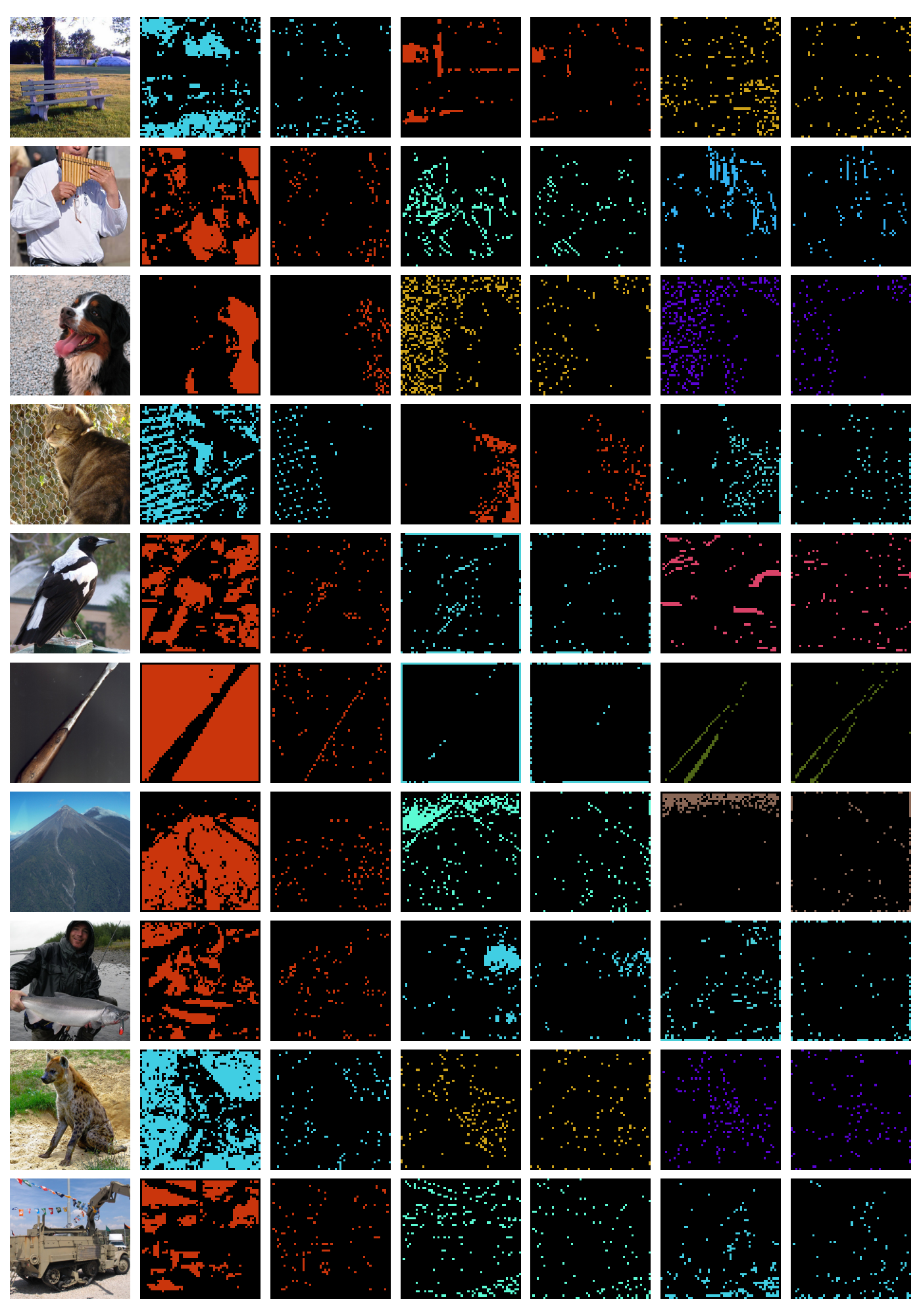}
    \caption{Visualization examples of query group and key selection from the first stage in DGT-B. The first column contains the original images. Every two adjacent columns is a query group and its corresponding key group for the remaining columns. %
    }
    \label{fig:visual}
\end{figure*}

Finally, we provide more visualization examples in Figure \ref{fig:visual}. These examples show that 1)different query groups attend to different keys according to their content. 2)The relevant keys are scattered everywhere rather than in a local area. These results show the advantages of our DG-Attention.

\bibliographystyle{named}
\bibliography{ijcai22}
\balance

\end{document}